\documentclass[10pt,conference]{IEEEtran}
\usepackage{cite}
\usepackage{amsmath,amssymb,amsfonts}
\usepackage{algorithmic}
\usepackage{graphicx}
\usepackage{longtable}
\usepackage{textcomp}
\usepackage{xcolor}
\usepackage[utf8]{inputenc}
\usepackage[english]{babel}
\usepackage{hyperref}

\begin{document}

\title{Graph Neural Network and NER-Based Text Summarization}

\author{\IEEEauthorblockN{Imaad Zaffar Khan}
\IEEEauthorblockA{\textit{NetId: izkhan2}\\
izkhan@illinois.edu}
\and
\IEEEauthorblockN{Amaan Aijaz Sheikh}
\IEEEauthorblockA{\textit{NetId: asheikh4}\\
asheikh4@illinois.edu}
\and
\IEEEauthorblockN{Utkarsh Sinha}
\IEEEauthorblockA{\textit{NetId: usinha3}\\
usinha3@illinois.edu}
}

\maketitle

\begin{abstract}
With the abundance of data and information in today’s time, it is nearly impossible for man, or, even machine, to go through all of the data line by line. What one usually does is to try to skim through the lines and retain the absolutely important information, that in a more formal term is called summarization. Text summarization is an important task that aims to compress lengthy documents or articles into shorter, coherent representations while preserving the core information and meaning. This project introduces an innovative approach to text summarization, leveraging the capabilities of Graph Neural Networks (GNNs) and Named Entity Recognition (NER) systems. GNNs, with their exceptional ability to capture and process the relational data inherent in textual information, are adept at understanding the complex structures within large documents. Meanwhile, NER systems contribute by identifying and emphasizing key entities, ensuring that the summarization process maintains a focus on the most critical aspects of the text. By integrating these two technologies, our method aims to enhances the efficiency of summarization and also tries to ensures a high degree relevance in the condensed content. This project, therefore, offers a promising direction for handling the ever-increasing volume of textual data in an information-saturated world.
\end{abstract}


\section{\textbf{Introduction}}
\subsection{\textbf{Background}}
\subsubsection{Introduction to Text Summarization}
Text summarization, in its essence, is the process of extracting the most important information from a text to produce a concise version that conveys the core meaning while reducing size of the text. This technique is divided into two primary categories: extractive summarization and abstractive summarization. Extractive summarization focuses on selecting and compiling key segments directly from the source text, whereas abstractive summarization involves generating new text that conveys the main ideas in a condensed form. Our focus lies on the former, exploring the nuances of extractive text summarization and its advancements.

\subsubsection{\textbf{Relevance in Today's World}}
In the current digital era, we are overwhelmed with textual information. From online news articles and scientific journals to social media feeds and corporate reports, the sheer volume of text available poses a significant challenge for individuals seeking to grasp information efficiently. Text summarization emerges as a potential and important solution in this context, enabling quick and effective access to essential content. Its applications span various domains where timely and accurate information extraction is quite important.

\subsubsection{\textbf{Transition to GNN and NER in Text Summarization}}
While there have been various advancements in the field of text summarization, the incorporation of Graph Neural Networks (GNN) and Named Entity Recognition (NER) stands out. These technologies address the unique challenges posed by the volume and complexity of modern data. GNN, with its capacity to model relational data, and NER, effective in identifying and classifying named entities in text, together improve the extractive summarization process. This integration can not only improves the accuracy of summarization but also ensure relevance and context-awareness, a critical aspect when dealing with diverse and extensive datasets.

\subsection{\textbf{Problem Statement}}
In the current times, we are confronted with an ever-growing size of textual data of great variety. This information overload not only challenges our capacity for comprehension but also strains traditional text summarization methods. These conventional approaches often fall short in addressing the complexity and scale of the data, leading to inefficiencies and missed insights.

Amidst this, Large Language Models (LLMs) have emerged as a powerful tool in text processing, showcasing an impressive ability to understand and generate human-like text. However, their implementation has the downside of significant computational costs and resource demands, making them less feasible for continuous use in various resource-constrained environments. This situation presents a critical need for more accessible and efficient text summarization solutions.

In this context, the integration of Graph Neural Networks (GNN) and Named Entity Recognition (NER) holds immense potential for revolutionizing extractive text summarization. GNNs excel in modeling complex data relationships, while NER systems are proficient in identifying and categorizing key entities within texts. Individually, they have shown promise in various text processing applications, but their combined use in extractive text summarization remains an uncharted territory.

This convergence presents a unique opportunity to address the limitations of current summarization methods, especially in managing large volumes of data efficiently and effectively. The core challenge lies in investigating the feasibility and effectiveness of combining GNN and NER in a unified framework. Such an integration aims not only to enhance the capacity for handling large-scale data but also to provide a resource-efficient alternative to the more demanding LLMs.

\subsection{\textbf{Objectives}}
The primary objectives of this project are outlined as follows:

\begin{enumerate}
    \item \textbf{Explore the Integration of GNN and NER for Extractive Text Summarization:} Implement the integration and test if in reality, the theoretical idea of the integration is practically effective once poised with the task of summariaztion. 

    \item \textbf{Create a Resource-Efficient Summarization Tool:} Design a text summarization system that operates with significantly lower computational resources compared to Large Language Models (LLMs), making advanced text summarization more accessible in various environments.

    \item \textbf{Evaluate the Performance of the GNN-NER Framework:} Conduct thorough testing and evaluation of the proposed GNN-NER framework across diverse real-world scenarios, assessing accuracy and efficiency.

    \item \textbf{Enhance Context-Awareness in Summarization:} Focus on improving the context-awareness in the summarization process, ensuring the summarized content retains the essential context and meaning of the original text with the help of proposed NER integration.
\end{enumerate}

\section{\textbf{Related Work}}
The field of extractive text summarization has seen significant advancements in recent years, with researchers exploring various methodologies to enhance the efficiency and accuracy of summarization processes.
\begin{enumerate}
    \item \textbf{Graph-Based Approaches}:

\cite{b1}Kadriua and Obradovica (2021) explored an extractive text summarization method using graphs, which offers insights into the structural analysis of texts for summarization purposes.
Further advancements in graph-based methods can be seen in other works, where the focus is on integrating graph structures with neural network models to capture contextual relationships in text.

\item \textbf{Named Entity Recognition in Summarization}:

\cite{b2}Alshibly et al. (2023) investigated text summarization of news articles using Named Entity Recognition (NER) via the SpaCy library, underscoring the importance of identifying key entities in text for effective summarization.

This is complemented by studies which delve into the integration of NER with machine learning algorithms to enhance the precision of extractive summarization.

\item \textbf{Language-Independent Summarization Techniques}:

\cite{b3}Hernández-Castañeda et al. (2022) discussed a language-independent approach to extractive automatic text summarization, focusing on automatic keyword extraction, which is crucial for understanding multi-lingual and cross-lingual summarization techniques.

\item \textbf{Algorithmic Innovations in Summarization}:

\cite{b4}P.Verma, A.Verma, and Pal (2022) presented an approach using fuzzy evolutionary and clustering algorithms for extractive text summarization, highlighting the role of advanced computational methods in this domain.
The evolution of algorithmic approaches can further be seen in various other studies, which explore the use of deep learning techniques for enhancing the adaptability and accuracy of summarization tools.
\end{enumerate}

Each of these studies contributes to a refined understanding of extractive text summarization. However, there remains a gap in exploring the combined potential of GNN and NER for this purpose. While individual studies have shown the efficacy of GNN in understanding complex relational data and NER in accurately identifying key entities, their integration in the context of extractive text summarization remains under-explored. This gap presents an opportunity for novel research, potentially leading to efficient summarization tools capable of handling the vast and varied nature of today’s textual data.

\section{\textbf{Methodology}}

This section outlines the methodology adopted for integrating Graph Neural Networks (GNN) and Named Entity Recognition (NER) in extractive text summarization. The methodology is structured into several key phases:
\subsection{\textbf{Proposed Architecture}}
The proposed architecture for our approach can be seen below:
\begin{figure}[h]
  \centering
  \includegraphics[width=0.5\textwidth]{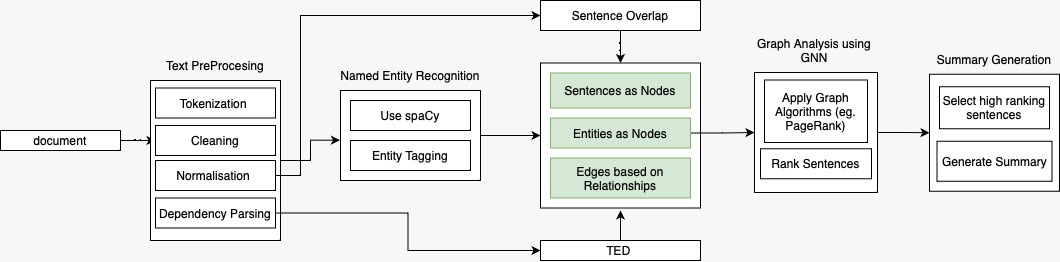}
  \caption{Architecture of the GNN and NER based text summarization system}
  \label{fig:architecture-diagram}
\end{figure}

The architecture diagram (Figure 1) illustrates the systematic flow of the proposed integration. The process is initiated with a document that undergoes a series of text pre-processing steps. These steps include tokenization, where the text is split into individual tokens, cleaning, to remove any irrelevant characters or spaces, normalization, which standardizes the text, and dependency parsing to analyze the grammatical structure of the text.

Following pre-processing, the text is passed through the NER phase, where spaCy is used for entity tagging, identifying named entities within the text. This phase is critical for recognizing important nouns and noun phrases that are essential to understanding the context of the document.

The extracted sentences and entities are then constructed into a graph, with sentences as nodes, entities also as nodes, and edges based on the relationships derived from the text and entity analysis. This graph is analyzed using GNN, specifically employing algorithms to evaluate the importance of each node within the graph structure.

In the final phase of Summary Generation, sentences are selected based on their ranking from the graph analysis. High-ranking sentences are deemed to carry significant informational weight and are chosen to be a part of the final summary. The summary is generated by strategically combining these sentences to reflect the main ideas of the original document while maintaining coherence and conciseness.

This architecture is designed to optimize the summarization process by leveraging the strengths of both GNN for capturing relational information and NER for emphasizing key entities, resulting in summaries that are both informative and representative of the source document.

\subsection{\textbf{Data Collection and Preprocessing}}
The project utilizes the CNN Daily Mail dataset, renowned for its diverse compilation of news stories, ideal for tasks in natural language processing, including summarization. The selection criteria for textual data included relevance to current events, diversity in topics, and rich entity representation to facilitate effective NER.

\begin{itemize}
    \item \textbf{Data Collection:} The dataset is accessed through established repositories, ensuring that the data is reliable and can serve its purpose. A brief table on the statistics of the CNN dataset can be seen below:
\begin{table}[h]
\centering
\caption{Overview of the CNNDailymail Dataset}
\label{tab:dataset-stats}
\begin{tabular}{|p{3.5cm}|p{4.5cm}|}
\hline
\textbf{Statistic}               & \textbf{Value}                  \\ \hline
Number of Articles               & $\sim$93,000(CNN)+$\sim$220,000(Daily Mail) \\ \hline
Number of Summaries              & Same as number of articles      \\ \hline
Average Length of Articles       & $\sim$800 (CNN), $\sim$1000 (Daily Mail) \\ \hline
Average Length of Summaries      & $\sim$55 (bullet points)        \\ \hline
Vocabulary Size                  & $\sim$350,000 to 400,000 unique tokens \\ \hline
Min Length of Articles           &  Generally around 100  \\ \hline
Max Length of Articles           & Varies, some exceed 2,000 words \\ \hline
Min Length of Summaries          & As low as 20  \\ \hline
Max Length of Summaries          & Can exceed 100 words            \\ \hline
\end{tabular}
\end{table}

The CNN Daily Mail dataset, can be seen from the statistics above, is a good choice for our project on extractive text summarization due to its substantial volume and diversity of length of articles as well as the diversity in the length of summaries provided. With over 300,000 combined articles from CNN and Daily Mail, the dataset offers a rich resource in terms of diversity and volume of data. The varying lengths of articles and summaries ensure that our system can adapt to a range of contexts and detail levels, which is an essential aspect for various use-cases. Moreover, the extensive vocabulary present in the dataset is advantageous for developing a robust model capable of understanding and summarizing a wide array of topics. This variability and complexity make the CNN Daily Mail dataset an ideal candidate to serve its purpose for our research.

    \item \textbf{Pre-Processing:} 

Data pre-processing forms the foundation of our text summarization model, as the subsequent graph representation relies heavily on the quality and richness of the input. The process comprises several stages, to refine the data for the construction of text-rich nodes and their edges/relationships within the graph. The steps are as follows:

\begin{itemize}
    \item \textbf{Tokenization:} The text is broken down into tokens. This step is crucial as it determines the granularity of the nodes in our graph. Each token potentially represents a node implying it to require careful segmentation.

    \item \textbf{Cleaning (Punctuation/Special Characters):} We remove punctuation and special characters that do not contribute to the semantic relationships in the graph. This step helps in reducing noise and focusing on the text that contributes to the meaning of the content.

    \item \textbf{Lemmatization:} Lemmatization is applied to reduce words to their base or dictionary form. We aim to capture the essence of the text in the graph, this step helps in generalizing the connections between nodes by reducing the words to their base form.

    \item \textbf{Lower Casing:} Converting all text to lower case ensures that the same words are not treated as different nodes due to case differences. 

    \item \textbf{Stop-Word Removal:} Common stop-words are filtered out as they usually do not carry significant meaning and are not useful as nodes in the graph. Their removal increases the text-richness of the remaining nodes, which bear more semantic weight.

    \item \textbf{Dependency Parsing:} Perhaps the most critical step, dependency parsing, is employed to understand the grammatical structure of the text, which informs the edges in our graph. It elucidates the syntactic relationships between tokens, allowing for a richer and more meaningful representation of the text as a graph.
\end{itemize}

Each of these pre-processing steps contributes to the construction of a graph where the nodes (words or entities) and their relationships (edges) are indicative of the content's semantic structure. Given that our approach heavily relies on the accurate representation of textual relationships through graphs, the pre-processing stage is not merely a prerequisite but a determinant of the summarization quality. The more text-rich and semantically accurate our graph, the better our model can identify and extract the most important/appropriate information for summarization. This attention to pre-processing detail ensures that our GNN can operate on data that closely mimics the interconnected nature of human language and thought.
\end{itemize}

\subsection{\textbf{Implementation of Named Entity Recognition (NER)}}
Named Entity Recognition (NER) is a critical stage in our text summarization process, going beyond basic pre-processing by enriching the text data with semantic information. This step acts as an additional input generation step. Output from the NER is given into the Graph Construction part of the pipeline. We utilize the spaCy library for its state-of-the-art NER capabilities.

An overview of this part of our pipeline can be seen below:

\begin{figure}[h]
  \centering
  \includegraphics[width=0.5\textwidth]{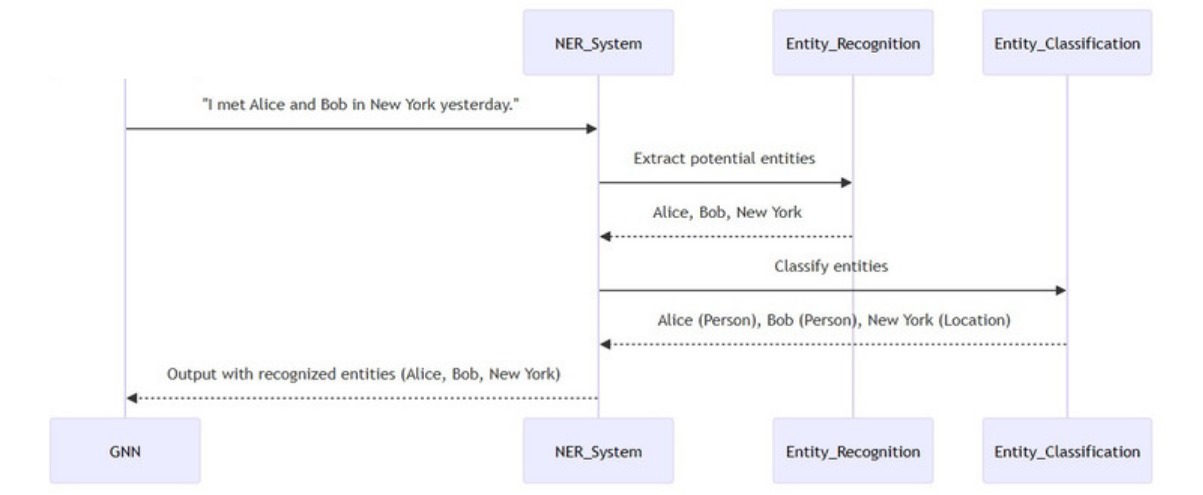}
  \caption{Architecture of NER Layer in Proposed Approach}
  \label{fig:architecture-diagram}
\end{figure}

Utilizing Figure 2, let us further look at the importance this layer:

\begin{itemize}
    \item \textbf{NER Model Selection:} We made use of spaCy's advanced pre-trained NER model, which identifies a wide range of entities with high precision. The model's robustness is derived from its training on a diverse and extensive corpus, enabling it to recognize and tag entities accurately across various domains\cite{b2}. Our selection of spaCy's model is based on its \textit{exceptional performance in benchmarks} and its ease of integration into our proposed pipeline.

    \item \textbf{Entity Identification:} The NER model tags text with entity types such as persons, organizations, locations, and others. This step is essential, as entities often carry significant weight in the meaning of a sentence and play a crucial role in summarization. By identifying these entities, we are able to preserve critical information that may otherwise be lost in a purely extractive approach.

    \item \textbf{Graph Construction with Entities:} In the subsequent stage of our pipeline, the recognized entities become nodes within our graph. This is not a simple transformation of text into graph form; rather, it is a careful transformation of the text into a structured representation that highlights the most meaningful components. The entities serve as pivotal points in the graph, often becoming central nodes that connect with various parts of the text, reflecting their importance in the narrative structure.

    \item \textbf{Semantic Enrichment:} The classification of entities facilitates a deeper understanding of the text by the GNN. It allows the GNN to not just process the text as a sequence of words but as a semantically rich network where the relationships between entities and other textual elements can be analyzed in a way that is closer to human cognition.

\end{itemize}

The implementation of NER is a decisive step that, in theory, should significantly influence the performance of our GNN model. By accurately identifying and classifying entities, we create a detailed semantic map of the text, which is essential for generating summaries that are both informative and reflective of the text's core themes.

\subsection{\textbf{Incorporation of Graph Neural Networks (GNN)}}
Graph Neural Networks (GNNs) are leveraged to capture the complex relational information among entities and contextual words in the text, effectively modeling the data as a graph. The process is visualized in Figure 3, which illustrates the transition from isolated nodes to a fully connected graph.

\begin{figure}[h]
  \centering
  \includegraphics[width=0.5\textwidth]{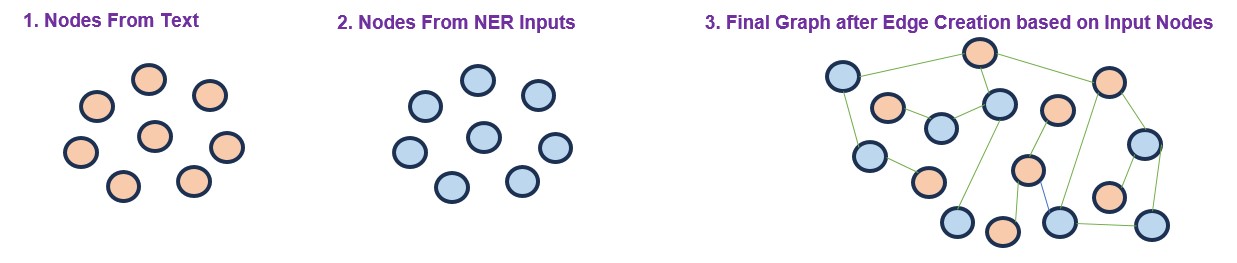}
  \caption{Progression of graph construction in GNN}
  \label{fig:gnn-process}
\end{figure}

Based on Figure 3, lets expand more below:

\begin{itemize}
    \item \textbf{GNN Model Selection:} Our selected GNN architecture is specifically designed to process text-based graph structures. It is designed to understand and make use of the relational dynamics between words and entities, which are essential for identifying crucial information in the text summarization process. The architecture is chosen for its ability to handle the heterogeneity of the graph's nodes and the complexity of their interrelations.

    \item \textbf{Graph Construction:} The graph is constructed in a two-step process. Initially, sentences and identified entities are treated as separate nodes (as shown in parts 1 and 2 of Figure 3). These nodes are then interconnected through edges that represent their relationships, which are determined by co-occurrence within sentences, syntactical dependencies, and semantic relatedness (part 3 of Figure 3). This results in a rich graph where the nodes are text elements and the edges represent the contextual and relational nuances of the original document.

    \item \textbf{Edge Formation:} Edges are not uniformly created; instead, their formation is informed by the depth of the relationship between nodes. Factors such as the frequency of co-occurrence and the strength of dependency links play a role in defining the edge weights, which are crucial for the GNN as it processes the graph to identify which nodes (text segments) are pivotal for the summary.

\end{itemize}

The incorporation of GNNs enables the system to move beyond linear text analysis, instead treating the text as a complex network, somewhat similar to human cognitive processes that interpret information. This networked approach allows for a more detailed understanding and generation of summaries, which are both comprehensive and contextually coherent.

\subsection{\textbf{Graph Analysis Using GNN}}
After graph construction, the next step is the graph analysis. This phase involves the application of graph-based algorithms to determine the centrality and importance of each node (sentence and entity) within the graph.

\begin{itemize}
    \item \textbf{Algorithm Selection:} For graph analysis, algorithms such as PageRank or other centrality measures are employed (discussed in detail in the following sub-section). These algorithms evaluate the importance of a node based on its connections, with the understanding that nodes with more connections or nodes connected to other highly connected nodes signify greater importance in the text.

    \item \textbf{Node Ranking:} Each node is scored and ranked according to the chosen algorithm. This ranking reflects the node's significance in the overall graph, which correlates to the importance of the corresponding sentence or entity in the context of the summary.
\end{itemize}

The outcome of this analysis is a list of sentences and entities ranked by their importance and relevance to the core content of the document, setting the stage for the final summarization.

\subsection{\textbf{Graph Analysis Algorithms}}
Graph analysis algorithms play an important role in interpreting the constructed graph and determining the importance of nodes for the summarization task. We have utilized and compared a variety of algorithms, each contributing a unique interpretation on node importance and network structure.

\begin{itemize}
    \item \textbf{PageRank:} PageRank is used to measure the importance of nodes in the graph. Nodes that are connected to many nodes, or to nodes that are themselves important, receive a higher score.

    \item \textbf{HITS (Hyperlink-Induced Topic Search):} HITS, focuses on identifying 'hubs' and 'authorities' within the graph\cite{b2}. A good hub represents a node that points to many authorities.A good authority is a node that is pointed at by many hubs.

    \item \textbf{Closeness Centrality:} This metric identifies nodes that are able to spread information efficiently through the graph. Nodes with a high closeness centrality can more effectively interact with all other nodes in the network.

    \item \textbf{Degree Centrality:} This is a measure of the number of connections a node has. In the context of summarization, nodes with high degree centrality have a greater likelihood to contain words or phrases common across the document, hence, pointing towards potential keywords or topics.

    \item \textbf{Betweenness Centrality:} This algorithm identifies nodes that serve as link between different parts of the graph. Nodes with high betweenness centrality can indicate sentences that relate various key concepts in the text.

    \item \textbf{Clusters:} Identifying clusters within the graph helps in understanding the modular structure of the text. Clusters can represent themes or topics within the document, aiding in the generation of a summary that covers a diverse range of points.
\end{itemize}

By utilizing and evaluating all the above mentioned algorithms, our goal is to try to select the algorithm that provided the best possible results.
Making use of these algorithms gives us the opportunity to in-depth compare our work, against previous works like \cite{b1},\cite{b2} and \cite{b3}, and actually visually verify if our approach of integrating the NER to our graph is fruitful or not.

\subsection{\textbf{Summary Generation}}

The final phase of our methodology is the generation of the summary. This phase capitalizes on the insights provided by the graph analysis algorithms, mentioned in the previous sub-section, to determine which parts of the text are most informative and are to be included in the summary.

\begin{itemize}
    \item \textbf{Sentence Selection:} Utilizing the rankings from graph analysis, sentences that are central to the text's meaning are selected. This selection ensures that the most relevant and important parts of the original document are preserved.

    \item \textbf{Redundancy Elimination:} To avoid repetitiveness, the algorithm identifies and removes redundant information. This ensures that the summary is not only concise but also non-repetitive.

\end{itemize}

The output is a summary that represents the key points and themes of the original text while being short and clear. This process mimics the human ability to identify and interpret the most useful information from a large text, providing users with a concise and informative version of the content.

\subsection{\textbf{Evaluation and Testing}}

Following the generation of summaries,evaluation and testing are conducted to ensure the effectiveness of the summarization pipeline.

\begin{itemize}
    \item \textbf{Evaluation Metrics:} We utilize standard evaluation metrics such as ROUGE to quantify the quality of the summaries by comparing them to reference summaries. These metrics assess the overlap of n-grams, word sequences, and word pairs between the generated and reference summaries.

    \item \textbf{User Studies:} Beyond automated metrics, user studies were conducted to gather qualitative feedback on the summaries' readability, informativeness, and overall utility.

    \item \textbf{Iterative Refinement:} Based on the evaluations, the summarization system undergoes iterative refinement to address any identified shortcomings, ensuring continuous improvement in performance.
\end{itemize}

In the next section, we look at the various challenges and pitfalls of evaluation techniques in the field of text summarization, followed by the overall results obtained by our proposed integration of NER in GNN.

\section{\textbf{Results}}

\subsection{\textbf{Evaluation Challenges in Text Summarization}}

Evaluation of text summarization systems presents unique challenges, mainly due to the subjective nature of what constitutes a \textit{good summary}. The lack of universally accepted evaluation metrics further complicates the assessment process. Traditional metrics tend to focus on surface-level text features which may not fully capture the semantic quality of a generated summary (an example of traditional metrics is word overlap).

\begin{itemize}
    \item \textbf{Semantic Richness:} Summaries need to be evaluated not just by the words they contain, but also by the semantic richness and the delivery of the main important ideas from the inputted text.
    \item \textbf{Diversity of Expression:} The same concept can be expressed in numerous ways, making word-to-word comparison not a sufficient measure for evaluation. A good summary might use different wording than the source text or reference summaries, leading to  low scores from automatic metrics.
\end{itemize}

These challenges highlight the need for more sophisticated evaluation approaches that can appreciate the nuanced task of summarization beyond mere lexical matching.

\subsection{\textbf{Limitations of Automatic Evaluation Metrics}}
Automatic evaluation metrics are often employed for their efficiency and objectivity. However, they inherently possess limitations, particularly when assessing the quality of text summaries. A more detailed view of the same can be seen below:

\begin{itemize}
    \item \textbf{Word-to-Word Comparison:} Metrics like precision and recall, while useful, may not fully recognize the quality of a summary if they only rely on direct word matches. Summaries containing synonymous expressions or paraphrases might be undervalued.
    
    \item \textbf{Potential for Low Scores:} Given the diversity of valid summaries for any given text, strict word-to-word comparison can result in lower scores for summaries that are, in reality, accurate and serve the required purpose, simply because they do not use the exact wording of the reference summaries.
\end{itemize}

Such limitations indicate the need for a balanced approach to evaluation, combining both automatic metrics and human judgment to capture the effectiveness of summarization systems more holistically.

\subsection{\textbf{ROUGE}}
The \textit{Recall-Oriented Understudy for Gisting Evaluation} (ROUGE) metric is widely used for evaluating automatic summarization and machine translation\cite{b3}.
We have a look below at the ROUGE metrics that we use:
\begin{itemize}
    \item \textbf{ROUGE-N:} This variant of ROUGE measures the overlap of n-grams between the generated summary and reference summaries. It is a valuable tool for assessing the presence of key phrases.
\end{itemize}

Despite its widespread adoption, ROUGE has its own limitations.  Therefore, while ROUGE scores can provide valuable insights, they should be interpreted in the context of these known constraints and supplemented with qualitative analyses to ensure a comprehensive evaluation of summarization performance.

In this section, we present the results of our text summarization system evaluated using the ROUGE metric, which is standard in the field of automatic text summarization. The results are presented in terms of both F1 scores, which combine precision and recall, and recall values alone, offering insight into the system's ability to capture the content of the reference summaries.

\subsection{\textbf{Findings}}
The following table presents the F1 scores for our approach across different ROUGE metrics.

\begin{table}[h]
\centering
\caption{ROUGE F1 Scores for Proposed Approach}
\label{tab:rouge-f1}
\begin{tabular}{|l|c|c|c|}
\hline
\textbf{Graph Algorithm} & \textbf{ROUGE 1} & \textbf{ROUGE 2} \\ \hline
PageRank            & 0.29         & 0.11             \\ \hline
HITS            & 0.30          & 0.11             \\ \hline
\textbf{Closeness}            & \textbf{0.30}          & \textbf{0.12}             \\ \hline
Betweenness            & 0.29          & 0.10             \\ \hline
Degree            & 0.30          & 0.11             \\ \hline
Clusters            & 0.28          & 0.09             \\ \hline

\end{tabular}
\end{table}

These scores reflect the system's effectiveness in producing summaries that are similar to the reference summaries, with a higher score indicating a more accurate summarization.

While F1 scores provide a harmonic mean of precision and recall, focusing on recall alone allows us to understand how well the system captures the entirety of the reference content.

\begin{table}[h]
\centering
\caption{ROUGE Recall Scores for Proposed Approach}
\label{tab:rouge-f1}
\begin{tabular}{|l|c|c|c|}
\hline
\textbf{Graph Algorithm} & \textbf{ROUGE 1} & \textbf{ROUGE 2} \\ \hline
PageRank            & 0.49         & 0.20             \\ \hline
HITS            & 0.42          & 0.17             \\ \hline
Closeness            & 0.46          & 0.18             \\ \hline
Betweenness            & 0.47          & 0.19             \\ \hline
Degree            & 0.48          & 0.20             \\ \hline
Clusters            & 0.42          & 0.15            \\ \hline

\end{tabular}
\end{table}

The recall scores are especially important in scenarios where the inclusion of all relevant information is critical, such as in comprehensive document summaries where missing details could lead to misinterpretation of the content.

\subsection{\textbf{Discussion}}
The obtained ROUGE scores suggest that our summarization system performs well in capturing the gist of the source documents. Again, it is to be noted that this statement is made on the grounds of the limitations of the overall evaluation metrics in the field of text summarization. Specifically, the F1 scores indicate a strong balance between precision and recall, while the recall scores suggest that the approach is quite effective at ensuring that the summaries are comprehensive. These findings are promising, showing that the novel approach can reliably generate summaries.

\section{\textbf{Comparative Analysis}}

\subsection{\textbf{Comparison Against Previous Works}}
A critical part of evaluating the effectiveness of our proposed summarization system is to compare its performance with that of existing approaches. We have selected several notable studies for comparison, with their results reported using the same ROUGE metrics on the same CNN/Dailymail Datset for consistency.

Below is a table that compares the best ROUGE F1 scores of our system with those of previous works of \cite{b1} and \cite{b2}. It is important to note that differences in test sets, pre-processing, and other experimental settings can affect comparability and hence provide an approxiamte comparison at the best.

\begin{table}[h]
\centering
\caption{Comparison of ROUGE F1 Scores with Previous Works}
\label{tab:comparison-f1}
\begin{tabular}{|l|c|c|c|}
\hline
\textbf{System} & \textbf{ROUGE-1} & \textbf{ROUGE-2} \\ \hline
Previous Work \cite{b1} & 0.28          & 0.11      \\ \hline
Previous Work \cite{b2} & 0.21          & - \\ \hline
\textbf{Our Approach}      & \textbf{0.30}          & \textbf{0.12}      \\ \hline
\end{tabular}
\end{table}

As illustrated in Table \ref{tab:comparison-f1}, our integrated approach demonstrates a better performance in comparison to the established benchmarks. This suggests that \textit{the integration of GNN and NER can indeed contribute to the improvement of extractive text summarization}.
\vspace{0.25cm}

Similarly, we compare the recall scores to assess the comprehensiveness of the summaries generated.

\begin{table}[h]
\centering
\caption{Comparison of ROUGE Recall Scores with Previous Works}
\label{tab:comparison-recall}
\begin{tabular}{|l|c|c|c|}
\hline
\textbf{System} & \textbf{ROUGE-1} & \textbf{ROUGE-2}  \\ \hline
Previous Work \cite{b1} & 0.48          & 0.17          \\ \hline
Our System      & 0.46       & 0.18                \\ \hline

\end{tabular}
\end{table}

The recall scores shown in Table \ref{tab:comparison-recall} highlight the system's ability to include relevant content in the summaries. Our system's performance in this area is indicative of its potential for practical applications where content coverage is crucial.

\subsection{\textbf{Comparison Against LLM's}}

Our approach, which integrates Graph Neural Networks (GNN) and Named Entity Recognition (NER), offers a different approach to that of LLM's, which we believe can be more cost effective. Here, we compare its performance against these LLMs to highlight the advantages and potential trade-offs.

The following table compares the summarization performance of our system against that of selected LLMs\cite{b5}. The comparison is based on the ROUGE metrics, considering both the F1 scores and recall.

\begin{table}[h]
\centering
\caption{Performance Comparison with LLMs}
\label{tab:llm-comparison}
\begin{tabular}{|l|c|c|c|}
\hline
\textbf{Model} & \textbf{ROUGE-1 F1} & \textbf{ROUGE-2 F1}   \\ \hline
BERTSUM+Transformer    & 0.43         & 0.20            \\ \hline
Our Approach      & 0.30          & 0.12   \\ \hline
\end{tabular}
\end{table}

This comparison indicates where our GNN-NER system stands in comparison to the current  LLMs. Particularly, it showcases the system's efficiency in generating summaries without the extensive computational resources typically required by LLMs.

One of the key advantages of our approach over LLMs is resource efficiency. LLMs, while powerful, require significant computational power, which can be a limiting factor in certain applications.

\subsection{\textbf{Case Study for effectiveness of Novel Approach}}

This case study aims to demonstrate the practical outcomes of different summarization methods. We compare the results of using Graph Neural Networks (GNNs) alone and in combination with a Named Entity Recognition (NER) based approach, and a Large Language Model (LLM-GPT) to provide a view on the summaries generated.

\subsubsection{\textbf{Method Description}}
The Various methods used:
\begin{itemize}
    \item \textbf{Integrated GNN-NER Approach:} Our proposed approach following the pipeline showcased in the above sections.
    \item \textbf{GNN-only Approach:} The process flow is similar to that of our proposed approach, apart from the additional input from NER block.
    \item \textbf{LLM-based Approach:}Using GPT to generate an extractive summary.
    
\end{itemize}

\subsubsection{\textbf{Input Text}}
\textit{"Cristiano Ronaldo and Lionel Messi will go head-to-head once more in the race to be this season's top scorer in the Champions League |xe2|x80|x93 although Luiz Adriano threatens to spoil the party. Both Barcelona and Real Madrid booked their spots in the semi-finals this week with victories over Paris Saint-Germain and Atletico Madrid respectively. The planet's best footballers have scored eight times in Europe this season. But Shakhtar Donetsk|xe2|x80|x99s Adriano, courted by Arsenal and Liverpool, has netted on nine occasions this term. Cristiano Ronaldo, in action against Atletico Madrid on Wednesday evening, has scored eight goals in Europe. Lionel Messi also has eight goals in the Champions League this term; one fewer than Luiz Adriano. Ronaldo and Messi will both play at least two more times after Real Madrid and Barcelona reached the last four. Adriano, who moved to Donetsk in 2007, scored five against BATE Borsiov in the group stages. His performance that night made history, with the 27-year-old becoming only the second player to score five times in a Champions League game. The other was Messi for Barcelona against Bayer Leverkusen in 2012. He also scored the third quickest hat-trick in the competition's history (12 minutes) as the Ukrainian side, knocked out by Bayern Munich in the round of 16, racked up the biggest-ever half-time lead (6-0) in Europe's premier tournament. |xe2|x80|x98I am in a good moment of my career and we'll do what will be best for me and for the club,|xe2|x80|x99 said Adriano last month when quizzed over his future. Adriano, who netted five times against BATE Borisov in the group, has scored more goals than any other player in the Champions League... he is out of contract in December and could move to the Premier League. |xe2|x80|x98With my contract set to expire and many good performances, it'll be difficult to stay in Ukraine. |xe2|x80|x99 Arsenal have sent scouts to watch Adriano in recent months, while Liverpool are also keen on the Brazilian. His contract with Shakhtar Donetsk runs out at the end of the year. Ronaldo and Messi however, remain in pole-position to top the scoring charts with Barcelona and Real Madrid both in the hat for the two-legged semi-finals to be played next month. Of the teams still in the pot, Neymar and Luis Suarez of Barcelona, Real Madrid's Karim Benzema and former Manchester United and City striker Carlos Tevez, now plying his trade for Juventus, each have six goals. The draw for the last four will take place on Friday."}
\vspace{0.25cm}
\subsubsection{\textbf{Summarization Results}}
The summaries obtained from each method can be seen in the figures below:

\begin{figure}[h]
  \centering
  \includegraphics[width=0.4\textwidth]{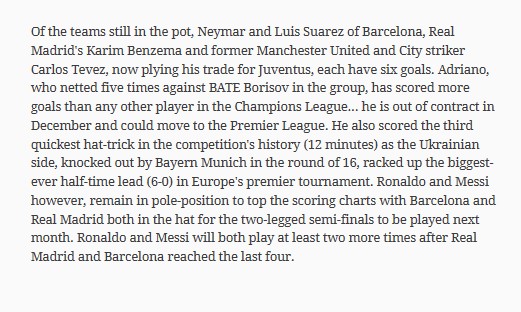}
  \caption{GNN+ NER Approach Generated Summary}
  \label{fig:architecture-diagram}
\end{figure}

\begin{figure}[h]
  \centering
  \includegraphics[width=0.4\textwidth]{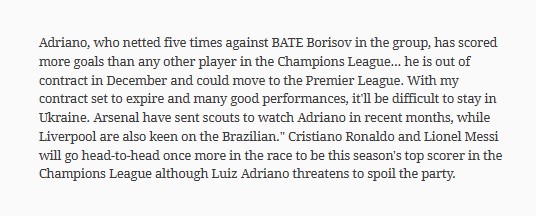}
  \caption{GNN Approach Generated Summary}
  \label{fig:architecture-diagram}
\end{figure}

\begin{figure}[h]
  \centering
  \includegraphics[width=0.4\textwidth]{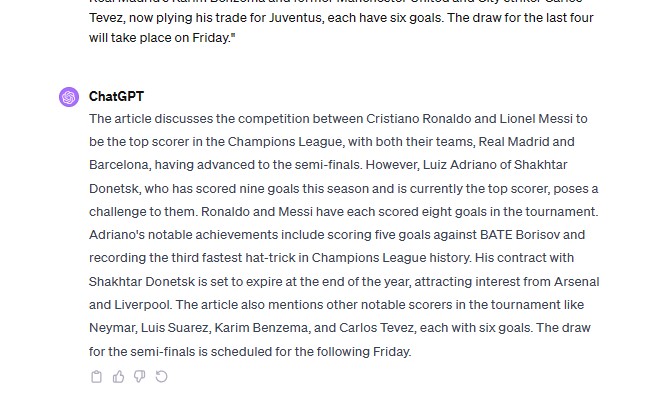}
  \caption{LLM(GPT) Approach Generated Summary}
  \label{fig:architecture-diagram}
\end{figure}

\subsection{\textbf{Analysis}}

This case study presents a comparative analysis of summaries generated using three different approaches. The comparison is based on the ability of the approaches for content coverage, coherence, conciseness, and focus on the main themes.

\begin{table}[h]
\centering
\caption{Comparative Analysis of Summarization Approaches}
\label{tab:summary-comparison}
\begin{tabular}{|p{1.5cm}|p{2cm}|p{2cm}|p{2cm}|}
\hline
\textbf{Aspect} & \textbf{GNN+NER} & \textbf{GNN Only} & \textbf{GPT} \\ \hline
\textbf{Content Coverage} & Covers relevant details & Focused on Adriano, No Clubs Mentioned. & Good Coverage. \\ \hline
\textbf{Coherence} & Moderate Coherency  & Less coherent & Good Coherency \\ \hline
\textbf{Conciseness} & Moderately concise & Less efficient organization of information. & Balanced  \\ \hline
\textbf{Focus on Main Themes} & Multiple themes covered &  Misses broader context  & Successfully captures the central themes \\ \hline
\end{tabular}
\end{table}

The table reveals that while GNN-based methods are effective in extracting key information, they may not always provide a coherent narrative. The GPT approach, in contrast, offers a more narrative and contextually rich summary, highlighting its strength in generating coherent and comprehensive summaries. However, based on human analysis, the ultimate question still lies on whether the computational expense is still worth it as the GNN+NER based approach generates a great quality summary nonetheless.

\section{\textbf{Conclusion}}

\subsection{\textbf{Summary}}
In this study, we presented a novel approach to extractive text summarization by integrating Graph Neural Networks (GNN) with Named Entity Recognition (NER). Our methodology leveraged the  GNN's ability to understand complex relationships within text and NER's precision in identifying relevant entities. Through a series of data pre-processing steps and the use of graph analysis algorithms, we constructed a  model that understands and summarizes large volumes of textual data with a degree of accuracy and semantic coherence. The approaches's performance, evaluated using the standard ROUGE metric, showcases its promising potential in generating concise summaries that retain the essential information of the original documents while its comparison with LLM's further validates its immense potential .

\subsection{\textbf{Future Work}}
The field of text summarization, particularly when combined with GNN and NER, is an uncharted territory with opportunities for further research. Future work may focus on several areas, including:
\begin{itemize}
    \item \textbf{Model Enhancement:} Refining the GNN architecture and exploring additional NER systems to improve the accuracy and richness of the summarization.
    \item \textbf{Evaluation Metrics:} Developing more sophisticated evaluation metrics that can capture the quality of summaries beyond word overlap, such as semantic fidelity and coherence.
    \item \textbf{Scalability:} Enhancing the model's scalability to handle datasets larger than CNN/Daily Mail, including real-time news feeds and extensive academic corpora.
\end{itemize}

\bibliographystyle{IEEEtran}

\end{document}